# Applying deep learning techniques on medical corpora from the World Wide Web: a prototypical system and evaluation

Jose Antonio Miñarro-Giménez [1,3], Oscar Marín-Alonso [1,2], Matthias Samwald [1*]

[1] Section for Medical Expert and Knowledge-Based Systems, Center for Medical Statistics, Informatics, and Intelligent Systems, Medical University of Vienna, Vienna, Austria.

[2] Dept. of Computer Technology, University of Alicante, Alicante, Spain.

[3] Institute of Medical Informatics, Statistics, and Documentation, Medical University of Graz, Graz, Austria.

[*] Corresponding author
E-mail: matthias.samwald [at] meduniwien.ac.at (MS)

# Abstract

**Background:** The amount of biomedical literature is rapidly growing and it is becoming increasingly difficult to keep manually curated knowledge bases and ontologies up-to-date. In this study we applied the word2vec deep learning toolkit to medical corpora to test its potential for identifying relationships from unstructured text. We evaluated the efficiency of word2vec in identifying properties of pharmaceuticals based on mid-sized, unstructured medical text corpora available on the web. Properties included relationships to diseases ('may treat') or physiological processes ('has physiological effect'). We compared the relationships identified by word2vec with manually curated information from the National Drug File – Reference Terminology (NDF-RT) ontology as a gold standard. We used different word2vec parameter settings and models to compare their impact on result quality.

**Results:** Our results revealed a maximum accuracy of 49.28% which suggests a limited ability of word2vec to capture linguistic regularities on the collected medical corpora compared with other published results. We were able to document the influence of different parameter settings on result accuracy and found and unexpected trade-off between ranking quality and accuracy. Pre-processing corpora to reduce syntactic variability proved to be a good strategy for increasing the utility of the trained vector models.

**Conclusions:** Word2vec is a very efficient implementation for computing vector representations and for its ability to identify relationships in textual data without any prior domain knowledge. We found that the ranking and retrieved results generated by word2vec were not of sufficient quality for automatic population of knowledge bases and ontologies, but could serve as a starting point for further manual curation. Future research should focus on how to combine word2vec tools with knowledge-based resources such as biomedical ontologies to create hybrid systems with greater accuracy and flexibility than either approach on its own.

*Keywords:deep learning; machine learning; text extraction; medicine; information retrieval; ontologies; software evaluation*

# Introduction

The large amount of biomedical information in databases such as PubMed [1] is a valuable source for automated information extraction [2] that facilitates the development of more efficient biomedical information retrieval systems. The concept of 'deep learning' has recently gained a lot of attention. It refers to unsupervised learning algorithms which automatically discover data without the need of supplying specific domain knowledge [3]. This approach usually has higher performance rates than supervised and informed methods when processing large unstructured corpora. However, the utility of these algorithms applied to realistic, domain-specific use-cases still needs further evaluation.

Word2vec [4] implements an efficient deep learning algorithm for computing high-dimensional vector representations of words and their relationships [5] based on unstructured text data. Once a vector model is created from a text corpus, word2vec provides two basic tools to use these models: *distance* and *analogy*. The *distance* tool provides a list of words closely related to a particular word from the vector model. These results also contain the corresponding cosine similarity of each related word that indicates how close the words are in the vector space model. The *analogy* tool, on the other hand, is able to query for textual regularities captured in the vector model through simple vector subtraction and addition.

For example, let us assume that we use word2vec to create a vector model of the words appearing in a large corpus of news articles. If the resulting vector space representation of cities and countries is projected to a two-dimensional representation, we can observe patterns such as those sketched in Figure 1. Not only are 'similar' (e.g., bordering) countries close to each other in vector space, we also see that their capital cities are arranged at predictable distances from their countries, because the deep learning algorithm was able to capture a notion of the 'capital city' relation between two entities from unstructured text sources. Because of this, the vector operation *Paris – France + Berlin* would result in a position in the vector space model that is close to the position of the word *Germany,* i.e., the regularities in the vector representation can be used to search for words related through a certain relationship to a query word. Because queries are simply defined 'by example', word2vec allows querying for poorly formalized relationships, which is of special interest in complex and evolving knowledge domains.

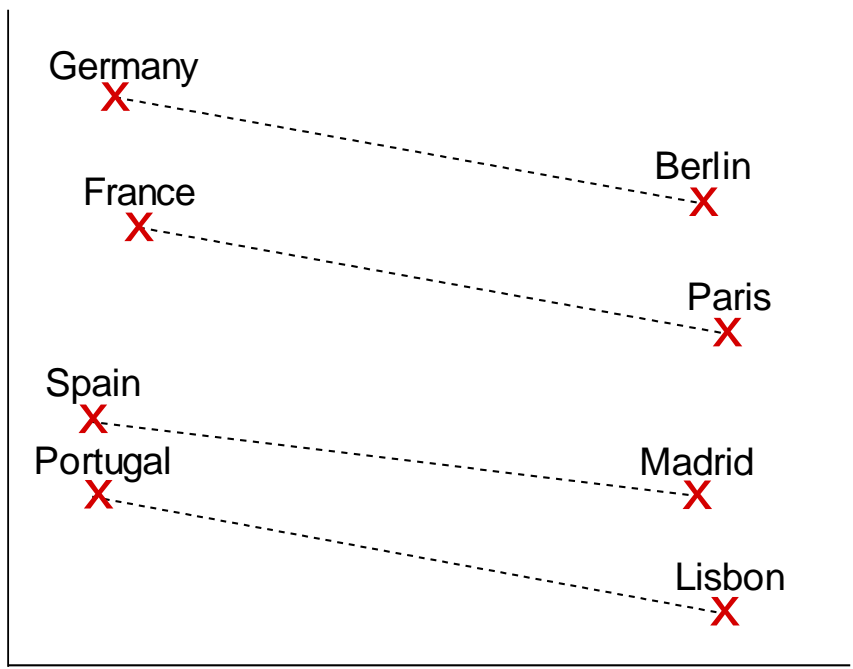

**Figure 1. Simplified, two-dimensional example of the regularities in the vector space representation generated by word2vec.**

These characteristics make word2vec of potential interest for improving the accessibility of unstructured medical content. For example, word2vec could assist the curation of structure knowledge bases and ontologies, and could help in refining information retrieval algorithms. However, deep learning algorithms such as word2vec are known to require very large amounts of training data to provide good results, and the amount of accessible, high-quality literature in specific domains (such as medicine) is often restricted, potentially decreasing the practical utility of the approach.

While similar approaches such as GloVe [6] were recently claimed to outperform word2vec tools for the unsupervised learning of word representations and word analogy, we utilized word2vec because it has been widely studied [6-8] and, therefore, our results can be easily compared with others.

In this paper we report on our evaluation of word2vec for clinically relevant medical content based on diverse, openly available, mid-sized medical text corpora. We compare the word relationships learned by word2vec with curated medical relationship encoded in the National Drug File – Reference Terminology (NDF-RT) ontology [9] to evaluate the results. The results of this exploratory work are intended to serve as an initial guidance that informs more in-depth work on applying word2vec in the medical domain.

# Material and Methods

## Word2vec

Word2vec is an efficient implementation of deep learning techniques based on two architectures, *continuous bag-of-words* (*CBOW*) and *skip-gram* (*SG*) [5], for computing continuous distributed vector representation of words from large datasets (up to hundreds of billions of words).

Word2vec requires training the corpora using one of these architectures. The training tool provides the following options: (i) type of architecture: continuous bag-of-words or skip-gram; (ii) the dimensionality of the vector space; (iii) the size of the context window in number of words; (iv) the training algorithm: hierarchical softmax and / or negative sampling; (v) the threshold for downsampling the frequent words; (vi) the number of threads to use; and (vii) the format of the output word vector file (an example command line is shown in Table 1).

**Table 1. An example of the parameters used by the word2vec training tool to generate the vector model of the medical corpora**.

| Command line parameters |
|---|
| word2vec *-train* corpora.txt *-output* vector-model.bin *-cbow* 0 *-size* 200 *-window* 5 *-negative* 0 *-hs* 1 *-sample* 1e-3 *-threads* 12 *-binary* 1 |

In this example, the training tool uses the corpus file "corpora.txt" to generate the vector model and serialize it to the file "vector-model.bin". It uses the skip-gram architecture (-cbow 0) with a vector dimension of 200 (-size 200) and a window size of 5 (-window 5). Besides, the vector model was generated using only the hierarchical softmax training algorithm (-hs 1) with a threshold of 0,001 for downsampling the frequent words (-sample 1e-3). The training tool uses 12 execution threads to generate the vector model as a binary file.

## Medical text corpora

We assembled a collection of openly available text repositories relevant to clinical medicine (excluding veterinary medicine) for use in this evaluation.

Two corpora were derived from PubMed. The first corpus was made up of PubMed abstracts with clinical relevance. To select abstracts of clinical relevance, a PubMed query was assembled by merging the lists of journals screened by the evidence-based medicine repositories DynaMed [10] and EvidenceUpdates [11]. The list was further manually edited, and additional constraints were added (e.g., excluding articles published before January 2005, excluding editorials) to create the final selection of PubMed abstracts. From this corpus of PubMed abstracts we derived another corpus by extracting the conclusion sections of the abstracts. The conclusion sections were further processed by expanding locally defined abbreviations in each abstract. This resulted in a smaller corpus made up of very high-quality, self-contained key assertions made in the clinically relevant research literature.

We created a corpus of medically relevant content from Wikipedia by selecting all articles that were associated with Wikiproject Medicine or Wikiproject Pharmacology [12] through manual curation of Wikipedia editors. This produced a corpus of Wikipedia articles with a good coverage of all major clinically relevant topics.

We also included two popular publicly available websites with content for medical professionals: Medscape [13] and Merck Manual [14]. We created a script for crawling medical content from these websites based on the PHPcrawl open-source library [15]. We also created scripts for stripping non-relevant portions of web pages (such as headers and footers) that would have significantly degraded the quality of the corpora. HTML markup was removed from the source data to yield raw text representations of the page contents.

Statistics on word counts and vocabulary sizes of the corpora generated in this way are summarized in Table 2.

**Table 2. Corpora used in the experiment.**

| Corpus | Word count | Vocabulary size |
|---|---|---|
| Clinically relevant subset of PubMed, full abstracts | 161,428,286 | 204,096 |
| Clinically relevant subset of PubMed, extracted key assertions (*'pubmed_key_assertions'*) | 17,342,158 | 47,703 |
| Merck Manual | 12,667,064 | 49,174 |
| Medscape | 25,854,998 | 63,600 |
| Clinically relevant subset of Wikipedia (*'Wikipedia'*) | 10,945,677 | 65,875 |
| Combined corpus (including all corpora above, *'combined'*) | 236,835,672 | 261,353 |

Word counts refer to the final corpora that were derived from source datasets after all processing steps. Vocabulary sizes refer to the number of distinct words found in each corpus. Underscored corpora were used for evaluating word2vec.

# National Drug File – Reference Terminology (NDF-RT) ontology

The NDF-RT is a formal representation of knowledge about drugs and is maintained by the US Department of Veterans Affairs. We chose the NDF-RT ontology as a reference for evaluating the results produced by the word2vec algorithms because it is one of the richest manually curated and openly available knowledge bases on medical drugs available. Several relationships between entities - such as 'may treat' or 'has mechanism of action' - were extracted from NDR-RT through SPARQL queries. The relationships that were extracted in this manner are described in Table 3.

**Table 3. Relationships between entities extracted from NDF-RT.**

| NDF-RT relationship | Number of distinct drugs annotated with at least one such relationship | Number of relationships |
|---|---|---|
| *'may_treat'* (drug X may treat disease / symptom Y) | 2,016 | 11,999 |
| *'may_prevent'* (drug X may prevent disease / condition Y | 781 | 1,909 |
| *'has_MoA'* (drug X has mechanism of action Y) | 2,436 | 9,126 |
| *'has_PE'* (drug X has physiological effect Y) | 2,250 | 13,769 |

In the text below we refer to relationships captured by NDF-RT as 'correct' relationships. Of course, this is a simplification of reality, it might well be possible that not all relationships found in NDF-RT are factually correct, and it is also likely that there are factually correct relationships missing from the NDF-RT ontology.

# Evaluation

In order to conduct the pilot evaluation of word2vec tools for medical text, we defined the following workflow for analyzing and evaluating the results produced by the word2vec tools: (1) gathering and processing openly available medical text corpora; (2) training vector space representations of these corpora with word2vec; (3) comparing results from the word2vec *distance* and *analogy* tools to assess how well the models captured the medical relationships such as 'may treat' and 'may prevent' from the medical reference ontology ; (4) calculating statistics to evaluate the impact of different parameter configurations.

# Gathering and processing medical corpora

The texts gathered from the selected sources needed to be pre-processed before they could be used for training vector models, since word2vec has no built-in functionalities for term normalisation or dealing with punctuation. We found that unprocessed corpora contained an abundance of basic syntactic variations and punctuation that had a negative impact on how word2vec indexes the terms and, therefore, the quality of the resulting vector space models.

The processing of corpora involved:

1. Removing all punctuation signs and superfluous whitespace, e.g., to avoid that "diabetes." (with a full stop) and "diabetes" would be erroneously indexed as two separate terms.
2. Transforming all words to lower-case, e.g., to avoid that "Diabetes", "diabetes" and "DIABETES" would be indexed as three separate terms.
3. Merging multiword terms, e.g. The term "glucose metabolism disorder" was transformed to "glucose_metabolism_disorder", which could then be processed like a single word by the word2vec algorithm. As word2vec processes every word in the corpus as a term, multiword terms cannot be represented in vector space. Therefore, we created a dictionary of relevant medical multiword terms from the NDF-RT ontology. When one term was found in the corpus, the words of the term were concatenated using underscores

The processing workflow is depicted in Figure 2.

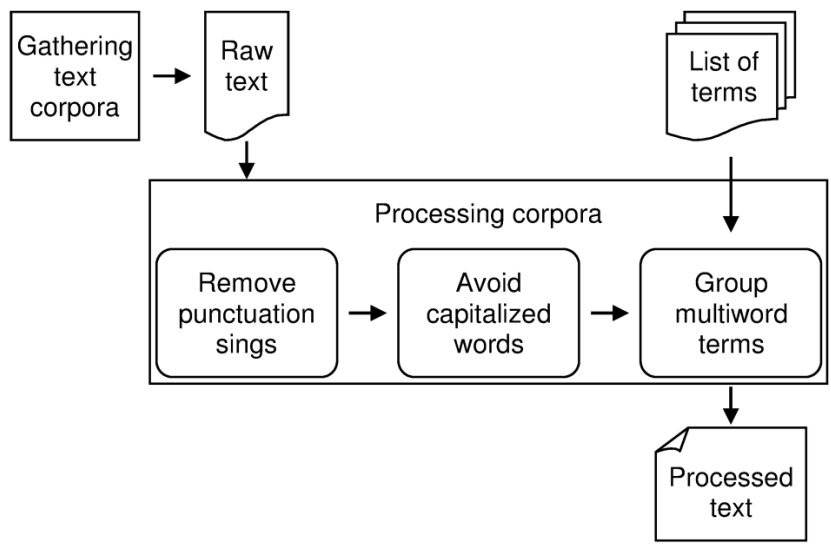

**Figure 2. Workflow for gathering and pre-processing the content of the corpora.**

# Training word2vec vector space models

The pre-processed corpora described above were used to build vector models with the word2vec training tool. The results of word2vec do not depend exclusively on the corpora but also on the parameters used. In order to test the impact of these parameter settings we compared the results of vector models trained with different parameter settings. We trained the corpora using (a) *CBOW* and *SG* vector model architectures; (b) 200, 300, 500 and 800 dimensionality of vector space; and (c) a word windows size of 5, 10 and 20. Other parameters, such as the training algorithm or the threshold for downsampling frequent words, were not varied to keep the complexity of results manageable.

# Assessment system

To assess the vector models of the trained corpora, the results of the word2vec *distance* and *analogy* tools were compared to the curated content of the NDF-RT ontology as a gold standard. To perform this evaluation, we developed a system that automatically queried a trained vector model using the *distance* and *analogy* tools of word2vec and matched the resulting list of words with the content of the NDF-RT ontology.

Figure 3 shows the software architecture of the assessment system. The "RESTful server" module made the word2vec *analogy* and *distance* tools accessible through RESTful services. RESTful services were deployed through the Java-based Jersey framework [16] and facilitated the access of the vector space models from external applications. Both services returned forty words with their cosine similarity values by default.

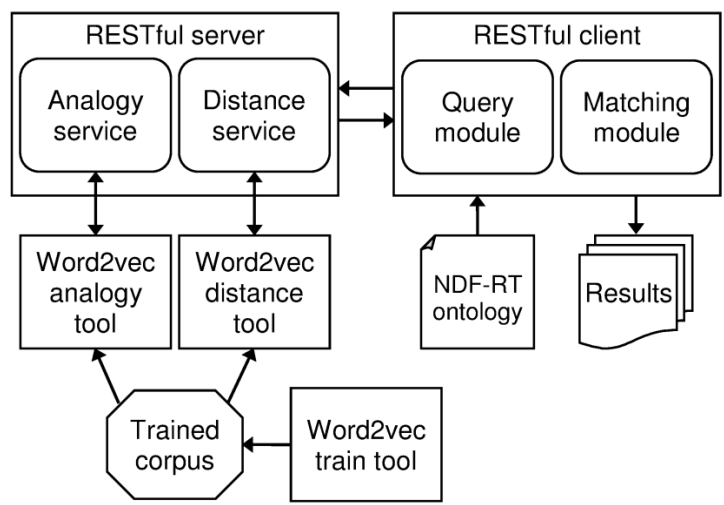

**Figure 3. The software architecture of the assessment system of word2vec tools using NDF-RT ontology.**

The "RESTful client" module (Figure 3) was responsible for gathering required information from the NDF-RT ontology, preparing the queries to consult the analogy and distance services, processing the responses and matching the retrieved terms with the information from NDF-RT to obtain the evaluation results. A batch process was also defined to automatically execute these tasks and to obtain the corresponding evaluation result for each trained corpus. The "Query module" collected subject-predicate-object triples for particular predicates from the NDF-RT ontology. Then, it created a list of unique subjects from all collected triples and used it to call the *analogy* and *distance* services. The "Matching module" received the results from the services and checked which words from the retrieved vector matched their corresponding words from the object values of the collected triples. Finally, the results of the word2vec tools evaluation were processed to calculate their accuracy values.

# Results and Discussion

The main results of this paper are: (1) the definition of a methodology to evaluate deep learning techniques in medical corpora; (2) the development of the batch system to automatically run the word2vec tools and match the results against the content of the NDF-RT ontology as a gold standard; and (3) the assessment of the results of the batch system for the evaluation of word2vec on medical corpora. The source code of the software for running these experiments is freely available for download at [17]; the software can be easily adapted to other use-cases, corpora and ontologies.

Removing sources of syntactic variability in the corpora proved to be a good strategy for increasing the utility of the trained vector models. As an example, Table 4 shows an excerpt of the result list when searching for words related to "aspirin" in the *combined* corpus with the *distance* tool. The results from the raw corpus include terms such as "[coumadin]" or "aspirin's" which contain irrelevant symbols. Furthermore, the lack of multiword terms in the raw corpus reduced the quality of retrieved terms, e.g., the term "acetylsalicylic_acid" (the active ingredient of Aspirin) is the closest to "aspirin" in the pre-processed corpus with a cosine similarity value of 0.83, whereas in the raw corpus the term "acetylsalicylic" is only found at the 5$^{th}$ position with a cosine similarity of 0.68.

**Table 4. Results of the Distance tool for the term "aspirin" when querying the pre-processed and raw versions of the combined corpus).**

| Rank | Pre-processed *combined* corpus | | Raw *combined* corpus | |
| --- | --- | --- | --- | --- |
| | Word | Cosine similarity | Word | Cosine similarity |
| 1 | acetylsalicylic_acid | 0.83 | clopidogrel | 0.79 |
| 2 | clopidogrel | 0.76 | ticlopidine | 0.72 |
| 3 | ticlopidine | 0.74 | thienopyridine | 0.69 |
| 4 | warfarin | 0.70 | warfarin | 0.69 |
| 5 | clopidogrel_bisulfate | 0.67 | acetylsalicylic | 0.68 |
| 27 | cilostazol | 0.60 | [coumadin] | 0.60 |
| 30 | antithrombotics | 0.60 | aspirin's | 0.59 |

The vector model based on the pre-processed corpus yields better results than the vector model based on the raw corpus, e.g., the term 'acetylsalicylic_acid' is the closest term to 'aspirin'.

The *analogy* tool aims to find textual regularities captured in the vector model to search for words related to a query word through a specific kind of relationship, while the *distance* tool simply returns words that are similar or generically related to a query word. To test whether the vector model actually captured textual regularities as expected, we tested if the analogy tool was superior in retrieving the right words from a specific relation. We calculated the accuracy of both tools for the three selected corpora: *combined*, *pubmed_key_assertions* and *Wikipedia* (see Table 2); and the relationships: *may_treat, may_prevent, has_PE* and *has_MoA* (see Table 3) from the reference ontology NDF-RT as a gold standard. Table 5 shows that the *analogy* tool indeed provided better results than the unselective *distance* tool, which underlines the ability of the *analogy* tool to recognize textual regularities beyond generic notions of relatedness. Models trained on the *combined* corpus, the largest corpus among the test corpora, produced best results, which confirms the hypothesis that word2vec model quality increases as corpus size increases. The number of correct results for the *may_treat* relationship with the *combined* corpus and the *analogy* tool was the best among all test cases with an accuracy 38.78%. Consequently, we choose the best configuration to run our evaluations regarding the influence of window-size and vector dimensionality parameters on the accuracy of the resulting vector models.

The optimisation of window-size and vector dimensionality parameters is described by the developers of word2vec as the most important ones for achieving good results. The window-size parameter corresponds to the span of words in the text that is taken into account, backwards and forwards, when iterating through the words during model training.

**Table 5. Accuracy values of the word2vec tools.**

| Corpus | Tool | may_treat | may_prevent | has_PE | has_MoA |
|---|---|---|---|---|---|
| combined | Analogy | **38.78**% | 13.82% | 2.54% | 8.14% |
| | Distance | 4.36% | 8.68% | 0.67% | 8.89% |
| pubmed_key_assertions | Analogy | 22.66% | 5.6% | 0.93% | 1.59% |
| | Distance | 2.66% | 5.85% | 0.37% | 3.94% |
| wikipedia | Analogy | 20.95% | 6.46% | 2.22% | 2.91% |
| | Distance | 1.39% | 3.56% | 0.32% | 3.56% |

Accuracy values of the word2vec tools using *combined, pubmed_key_assertions* and *Wikipedia* corpora, with *analogy* and *distance* tools, and the *may_treat, may_prevent, has_PE* and *has_MoA* relationships. The best accuracy was obtained with the combined corpus and analogy tool when querying *may_treat* information.

Figure 4 shows the accuracy when querying the resulting vector models generated with *SG* and *CBOW* architectures and with three different window sizes, 5, 10 and 20. Larger window sizes increase the time required to train vector models, so choosing the optimal window size can reduce computing time. Our findings show that the window-size parameter is also relevant for improving the accuracy of the gathered medical corpora. With our corpora, the optimal value is a window size of 10.

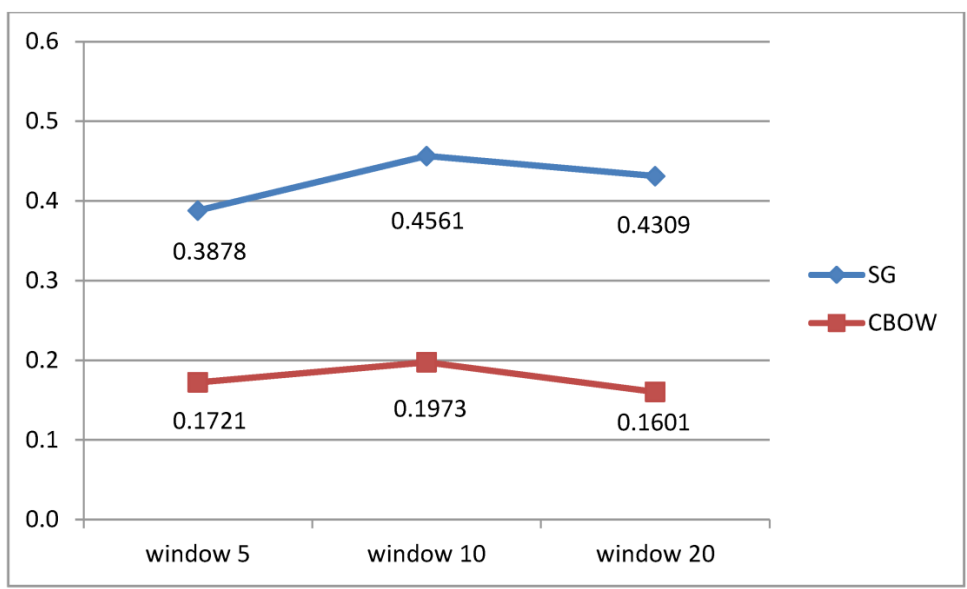

**Figure 4. Comparison between the accuracy values using skip-gram (SG) and continuous bag-of-words (CBOW) architectures. Comparisons were conducted for window sizes of 5, 10 and 20. Results are based on a 200-dimensional vector model, the *combined* corpus and the *may_treat* relationship from NDF-RT ontology as the gold-standard for querying the word2vec *analogy* tool.**

The influence of the vector dimensionality parameter was evaluated using vector models generated with 200, 300, 500 and 800 vector dimensions, and the *SG* and *CBOW* architectures. Figure 5 shows the comparison between the accuracy values with various parameter configurations. In our experiments, the best ranking results were obtained with the *SG* architecture and a vector dimensionality of 300, while worst ranking results were observed with the *CBOW* architecture and a vector dimensionality of 800. The results suggest that the *SG* architecture consistently yielded better result ranking than the *CBOW* architecture. This is also consistent with other evaluations such as [6] which obtained an accuracy of 61% with SG, a dimension vector of 300 and 1B word corpus. We observed a U-shaped relationship between the dimensionality of the model and the accuracy of results with *SG* and *CBOW* architecture, so a higher vector dimensionality will not provide better results than the stated ones. A higher dimension of the vectors implies a bigger size of the resulting vector model. Therefore, having a vector model with a dimension of 300 requires less memory space and provides better results than a vector model with 500 or 800 vector dimensions.

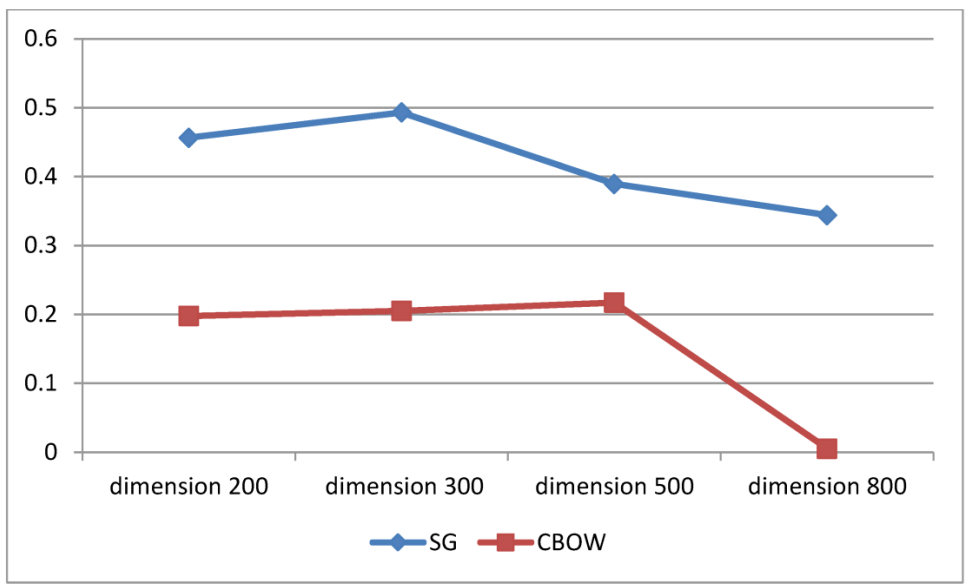

**Figure 5. Accuracy of the vector models generated using the *combined* corpus with skip-gram (SG) and Continuous bag-of-words (CBOW) architectures. Comparisons were conducted for vector dimensionalities of 200, 300, 500 and 800. The may_treat information from NDF-RT ontology is used as the gold standard for querying word2vec analogy tool.**

# Conclusions

The ability of word2vec to retrieve expected terms from the size-restricted corpora we used is not suitable for applications requiring high precision since we only obtained an accuracy of 49.28%. These modest results could be explained by the restricted size of gathered medical corpora as well as the complexity of the medical knowledge domain. It could be of interest for future research to test this tool with larger, commercially available medical corpora.

Due to the complexity of medical terminology, we found pre-processing of corpora necessary to decrease syntactic variability. We also found that many relevant medical terms of interest were composed of multiple words, and that ontology-based pre-processing of these terms led to a marked improvement of the results from the word2vec tools.

As expected, the *analogy* tool produced better results for identifying related entities for a specific type of relationship than the *distance* tool, and the largest corpus provided better results than smaller sub-corpora. As a consequence of our results, we recommend to use *SG* architecture rather than *CBOW* to train other medical datasets because such architecture always produced the best accuracy values. Moreover, the combination of a 10 window-size with a 300 vector dimension produced the best results among all tested parameters configurations. We also conclude that a vector dimension greater than 800 and a window size greater than 20 are not recommended due to the observed quick deterioration of accuracy values.

Regarding the low matching rate observed in our evaluation with mid-sized medical corpora, our future research will focus on how to improve accuracy values by combining word2vec analogy tool with knowledge-based resources such as ontologies to create hybrid systems and also running the evaluations with larger, commercially available corpora.

# Acknowledgments

We thank the word2vec team for providing assistance with tuning parameters of the word2vec tools.